\documentclass{article}

\usepackage[utf8]{inputenc}
\usepackage[T1]{fontenc}
\usepackage{hyperref}
\usepackage{url}
\usepackage{booktabs}
\usepackage{amsfonts}
\usepackage{amsmath}
\usepackage{nicefrac}
\usepackage{microtype}
\usepackage{graphicx}
\usepackage{xcolor}
\usepackage{amsfonts}
\usepackage{lipsum}
\usepackage{booktabs}
\usepackage{multirow}
\usepackage{graphicx}
\usepackage{subcaption}
\usepackage{array}
\usepackage{bbold}
\usepackage{float}
\usepackage[margin=0.95in]{geometry}

\title{
   Safe Navigation: Training Autonomous Vehicles using Deep Reinforcement Learning in CARLA\\
  \vspace{0.5em}
}

\author{
}
\date{}

\begin{document}

\maketitle
\vspace{-1.5cm}
\begin{center}
    Ghadi Nehme\footnote{Stanford University}, Tejas Y. Deo\footnotemark[\value{footnote}]
\end{center}
\vspace{0.5cm}

\begin{abstract}
   Autonomous vehicles have the potential to revolutionize transportation, but they must be able to navigate safely in traffic before they can be deployed on public roads. The goal of this project is to train autonomous vehicles to make decisions to navigate in uncertain environments using deep reinforcement learning techniques using the CARLA simulator. The simulator provides a realistic and urban environment for training and testing self-driving models. Deep Q-Networks (DQN) are used to predict driving actions. The study involves the integration of collision sensors, segmentation, and depth camera for better object detection and distance estimation. The model is tested on 4 different trajectories in presence of different types of 4-wheeled vehicles and pedestrians. The segmentation and depth cameras were utilized to ensure accurate localization of objects and distance measurement. Our proposed method successfully navigated the self-driving vehicle to its final destination with a high success rate without colliding with other vehicles, pedestrians, or going on the sidewalk. To ensure the optimal performance of our reinforcement learning (RL) models in navigating complex traffic scenarios, we implemented a pre-processing step to reduce the state space. This involved processing the images and sensor output before feeding them into the model. Despite significantly decreasing the state space, our approach yielded robust models that successfully navigated through traffic with high levels of safety and accuracy.
\end{abstract}


\section{Introduction}
The development of safe and reliable autonomous driving technology remains a critical research area, with numerous studies exploring various approaches to enhancing the safety of self-driving vehicles. When it comes to beyond level 4 in autonomous driving technology, the vehicle should be able to deal with various kinds of situations around it in order to arrive at the targeted destination successfully. Different sensor and algorithm-based approaches have been tried and developed to navigate safely to the target destination in the CARLA simulator \cite{terapaptommakol2022design} \cite{jeon2022carla} \cite{perez2022deep}. Conventional control systems are based on mathematical models, but they only control the vehicle in a limited range of situations. Therefore, Machine Learning (ML) algorithms have been applied in autonomous systems \cite{kiran2021deep} \cite{sanil2020deep} to better control vehicles in varied situations. 

\subsection{CARLA (Car Learning to Act) Simulator}
CARLA enables testing algorithms in safe and controlled environments without the risk of accidents and injuries. It provides a cost-effective and efficient way to test and improve autonomous driving algorithms before deploying them in real-world vehicles. It provides a realistic 3D environment that mimics real-world scenarios, including urban, suburban, and rural environments. It provides a platform for collecting and analyzing data generated during the simulation and training and improving autonomous driving algorithms \cite{jaafra2019seeking} \cite{dosovitskiy2017carla}.

\subsection{Deep Reinforcement Learning (Deep RL)}
Deep RL combines the power of Deep Neural Networks (DNN) with RL techniques. At the heart of Deep RL lies the Markov decision process (MDP), a mathematical framework used to model sequential decision-making problems. In Deep RL, MDPs are used to represent the agent's interaction with the environment, allowing it to learn optimal policies that maximize its long-term rewards. Deep RL enables agents to learn complex representations of the environment, allowing them to make informed decisions based on high-dimensional sensory input. A hierarchical architecture for sequential decision-making in autonomous driving using Deep RL is implemented in this paper \cite{moghadam2019hierarchical}. 

\begin{figure}[H]
\centering
\includegraphics[width=0.5\textwidth]{ 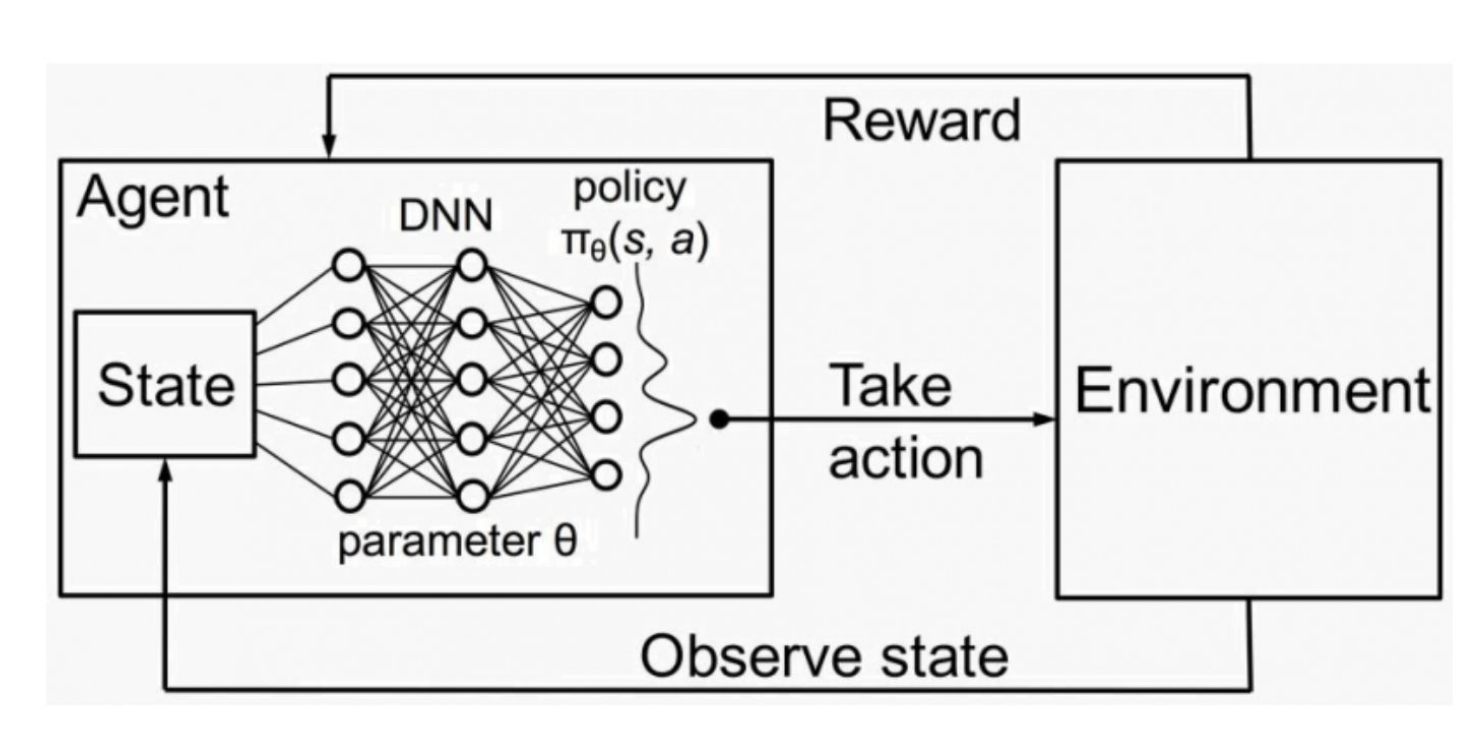}
    \caption{Deep RL Framework}
\end{figure}

\subsection{Deep Q-Network (DQN)}
DQNs are a neural network architecture used in RL that combines Q-learning with Deep Learning (DL) techniques. They approximate the Q-function rather than storing it in a lookup table, making them suitable for complex and high-dimensional environments. DQN's use of experience replay prevents overfitting and stabilizes training, making it a powerful method for learning a policy in a model-free approach. Once the DQN is trained, we get the optimal policy as follows:
$$\pi(s) = \textrm{arg} \max_a Q(s,a)$$

\section{Problem Statement} 

The deployment of autonomous vehicles on road to safely navigate in uncertain environments to the final destination require fast computation, low latency, and high accuracy. Therefore, there is a need to develop efficient techniques to process high-dimensional input data from sensors to reduce the state space and improve the training and testing efficiency. This paper proposes a Deep RL method using DQN, which includes sensor data pre-processing steps to reduce the state space and demonstrates its effectiveness in successfully navigating through 4 traffic scenarios with high levels of safety and accuracy.

\section{Implementation}

In this section, we describe the implementation of our approach for safely navigating a car in traffic from a starting position to a final destination in CARLA. The implementation is divided into several subsections that describe the key components of our approach.

\subsection{State and Action Space Representation}

\begin{figure}[H]
  \centering
  \begin{subfigure}[b]{0.39\textwidth}
    \includegraphics[width=\textwidth]{ 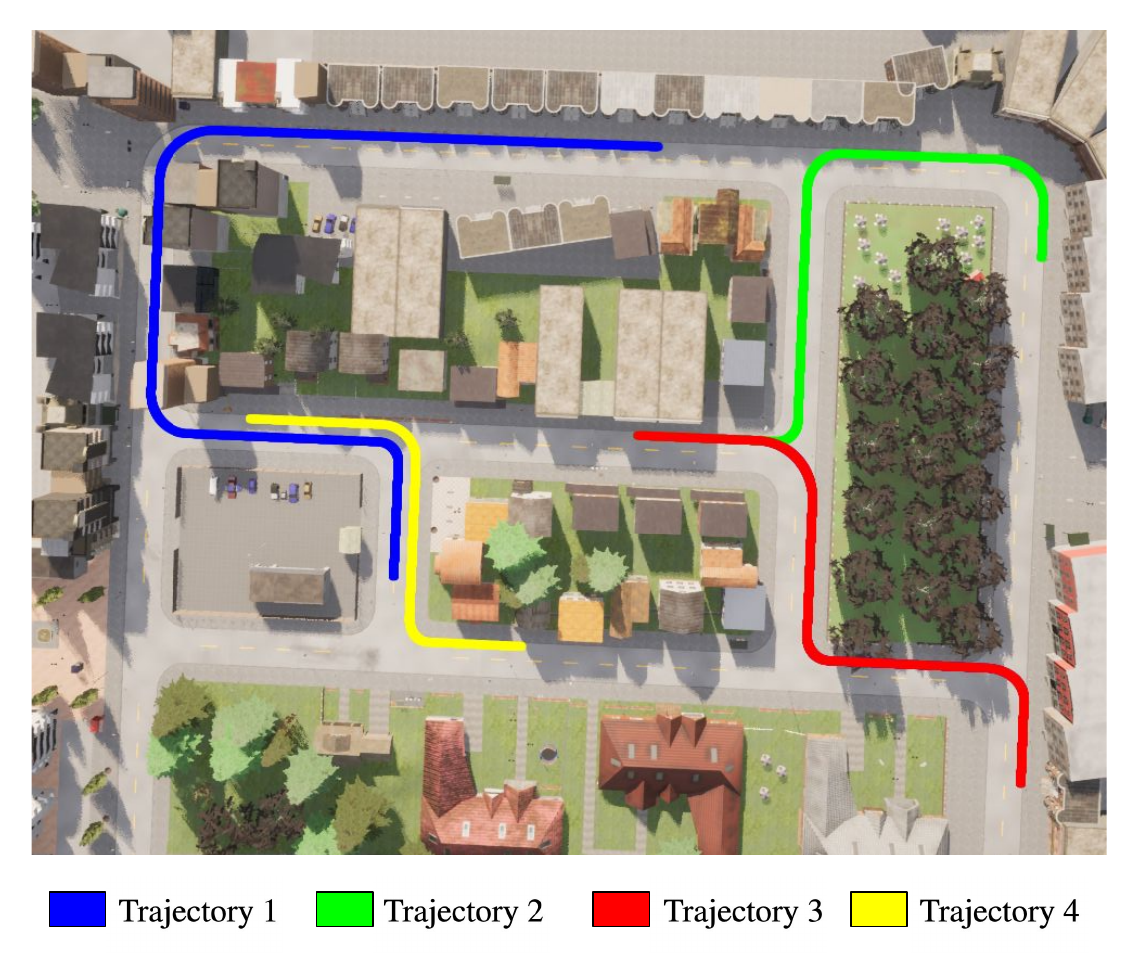}
    \caption{}
    \label{fig:traj}
  \end{subfigure}
  \hfill
  \begin{subfigure}[b]{0.55\textwidth}
    \includegraphics[width=\textwidth]{ 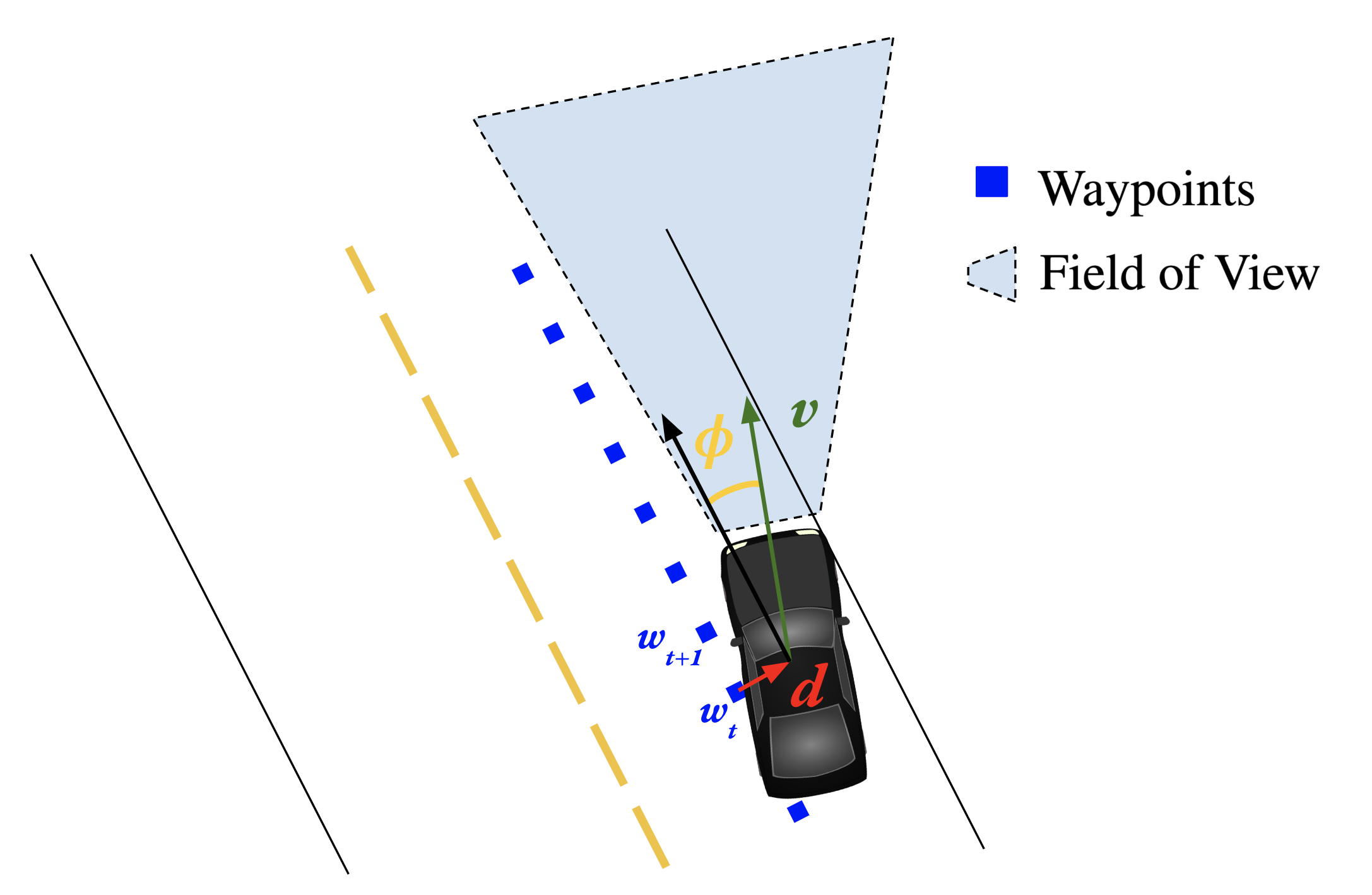}
    \caption{}
    \label{fig:scheme}
  \end{subfigure}
  \caption{(a) The trajectories using Waypoints in Carla between specific starting positions and final destinations, used for testing the driving and combined models, (b) Schematic showing the car, its Field of Vision, the waypoints of the path that the car should follow, the velocity $v$ of the car, the orientation of the ego vehicle with respect to the road $\phi$ and the signed distance to the centerline of the road $d$}
  \label{fig:main}
\end{figure}

To navigate a car safely in traffic in the CARLA simulation environment, we first start by building the path $\mathcal{P}$ that the car should follow between the starting position and the desired final destination using waypoints in CARLA (Figure \ref{fig:traj}). We then need to extract relevant information from the available sensors. We use the segmentation and depth images provided by the simulator to detect and locate obstacles, such as vehicles and pedestrians, to estimate their distance from the ego vehicle.

\begin{figure}[H]
\centering
\includegraphics[scale = 0.47]{ 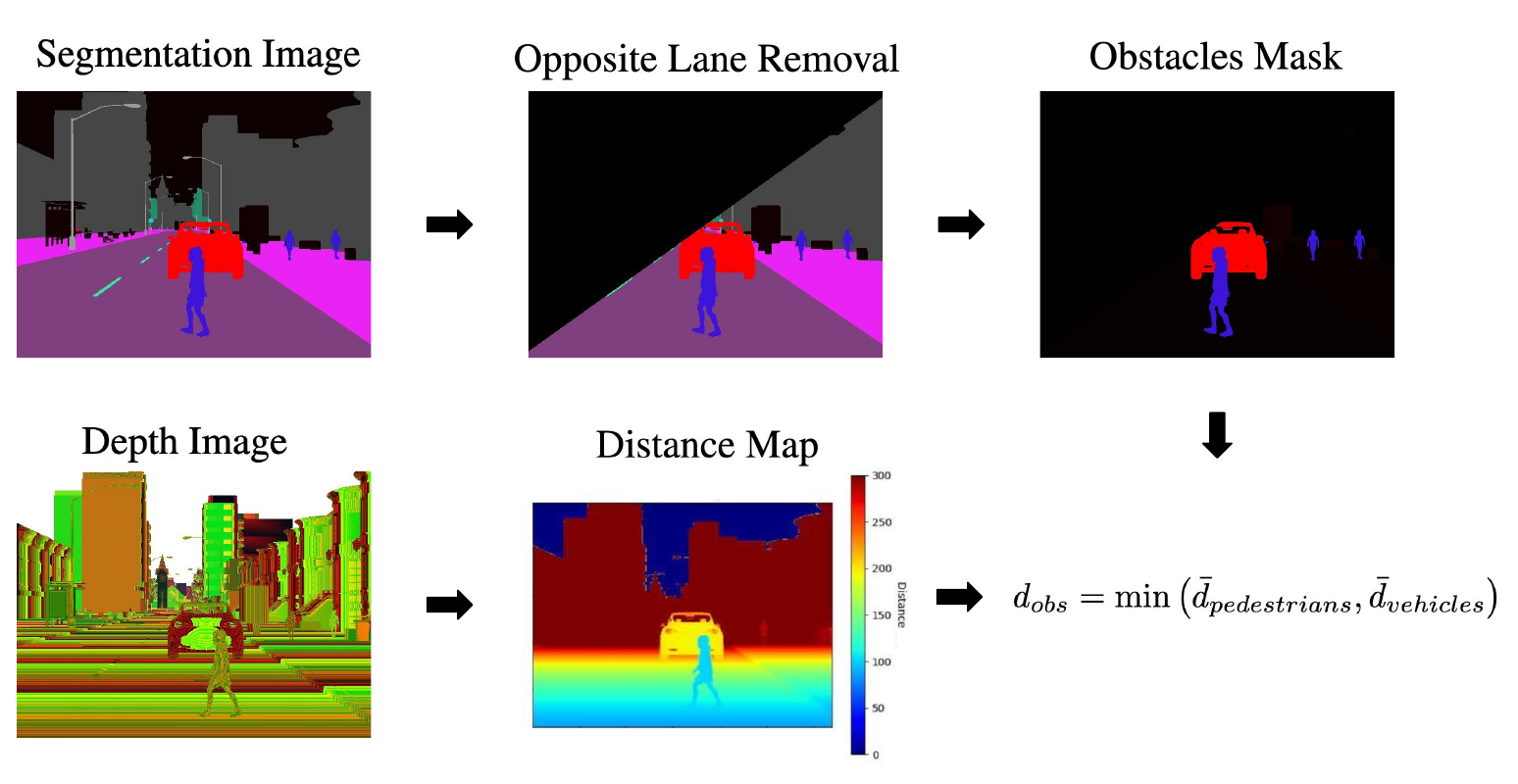}
    \caption{Pre-processing of Segmentation and Depth Images to compute the distance to the closest obstacle $d_{obs}$}
\label{prep}
\end{figure}

Specifically, we first pre-process the segmentation image to remove data from the opposing lane, where cars coming from that direction should not make our agent stop. To do that, we fit the road line using a polynomial of degree 2 and set all the values of the image on the left of that curve to zero (Figure \ref{prep}). Then, we extract the distance map from the depth image and detect the area in the processed segmentation image where we have vehicles and pedestrians. Finally, we extract binary masks of the obstacle regions from the pre-processed segmentation images, and then compute the mean distance value within these regions for each of the pedestrians and the vehicles. We then take the minimum of these distances as an estimate of the distance to the closest obstacle (Figure \ref{prep}).

To extract the relevant geometric information about the road and the ego vehicle, we use the planned trajectory $\mathcal{P}$ defined above. We compute the orientation of the ego vehicle with respect to the road $\phi$, by finding the closest waypoint $w_t$ on the trajectory and computing the relevant values based on the difference between the yaw angles of the road and the vehicle (Figure \ref{fig:scheme}). We also compute the signed lateral distance $d$ between the agent and the centerline of the road using the following formula:

$$d = ||v|| \sin \left(\textrm{sgn}(u\times v) \cos^{-1} \left(\frac{u\cdot v}{||u|| ||v||}\right)\right), $$

$\textrm{with } u = w_{t+1} - w_t, v = w_t - p \textrm{ and } p \textrm{ the position of the agent}$ (Figure \ref{fig:scheme}).

To reduce the complexity of the state space, we thus select four variables that capture the most salient aspects of the ego vehicle's position and velocity, as well as the presence and proximity of obstacles. These variables are the lateral distance to the centerline of the road ($d$), the orientation of the ego vehicle with respect to the road ($\phi$), the minimum distance to an obstacle ($d_{obs}$), and the velocity of the ego vehicle ($v$).

Concerning the action space in CARLA, we are able to select a throttle value between 0 and 1, a steer value between -1 and 1 and a brake value between 0 and 1. To discretize this action space, we choose 6 actions that would allow the agent to complete any trajectory given an optimal policy. The 6 actions are: brake (0), go straight (1), turn left (2), turn right (3), turn slightly left (4), turn slightly right (5).

\subsection{DQN Models}

We use Deep Q-Networks (DQNs) to learn a policy for the ego vehicle that can safely navigate through traffic. To this end, we train two separate DQN models, one for the braking action and one for the driving actions. During training, we use an epsilon-greedy policy to balance exploration and exploitation. The models are trained on 40 episodes using the Adam optimizer with a learning rate of 0.0001, a batch size of 16, and a discount factor of 0.99. We also use experience replay and target network techniques to stabilize the training process.

\subsection{Braking Model}

The braking DQN model takes the current values of $d_{obs}$ and $v$ as inputs and outputs Q-values for two actions: braking (0) or driving (1). The model architecture consists of one fully connected layer with 2 units, followed by a linear activation (Figure \ref{combined}). The model is trained on a straight road with a car spawned at a random distance from the agent. The goal is to make the agent stop close to the front car without colliding with it to succeed in this scenario. We thus chose a reward function that rewards the agent for high speeds far away from the car and low to zero speeds close to the car and penalizes the collisions. The reward function given to the agent is defined as follows:

\begin{align*}
    R_b (v,d_{obs},a) = & \left[v<10\, d_{obs}+10\right]\wedge \left( 3[a= 0]  -  [a= 1] \right) + 2 \left[v>10 \,d_{obs}+10 \right] \wedge \left(2\,[a= 1] - 1\right) \\
    &-10\,[v < 1] \wedge [d_{obs} > 100] +200 \,[v = 0] \wedge [d_{obs} < 150] -200 \,[\textrm{collisions} > 0]
\end{align*}

We train our agent for 40 episodes and observe that the average reward during training is increasing and then reaches a plateau around 100 (Figure \ref{fig:brak}). This shows that the car was able to learn when to brake based on $v$ and $d_{obs}$, while avoiding collisions and maximizing its speed far away from the front car.

\subsection{Driving Model}

\begin{figure}[H]
  \centering
  \begin{subfigure}[b]{0.49\textwidth}
    \includegraphics[width=\textwidth]{ 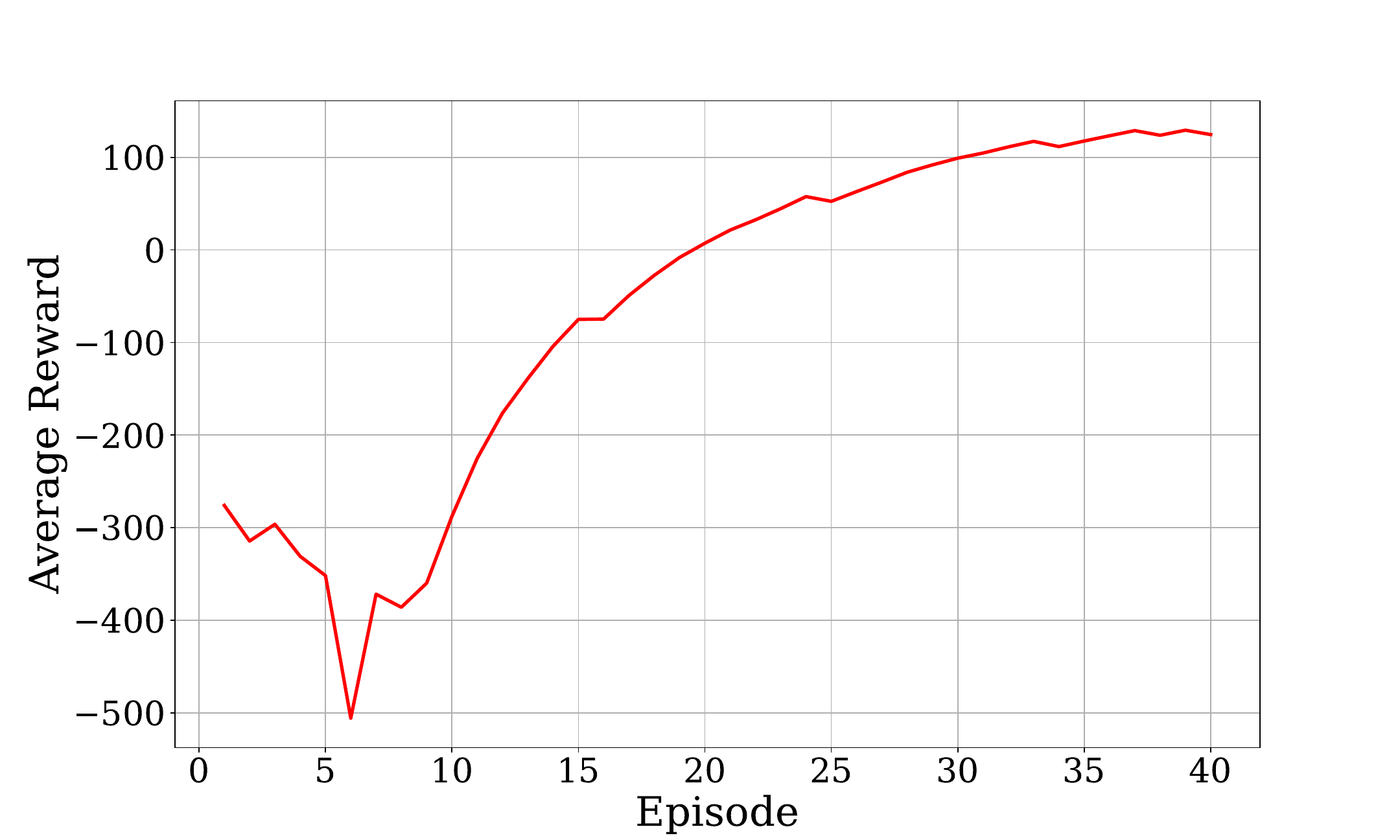}
    \caption{Braking Model}
    \label{fig:brak}
  \end{subfigure}
  \hfill
  \begin{subfigure}[b]{0.49\textwidth}
    \includegraphics[width=\textwidth]{ 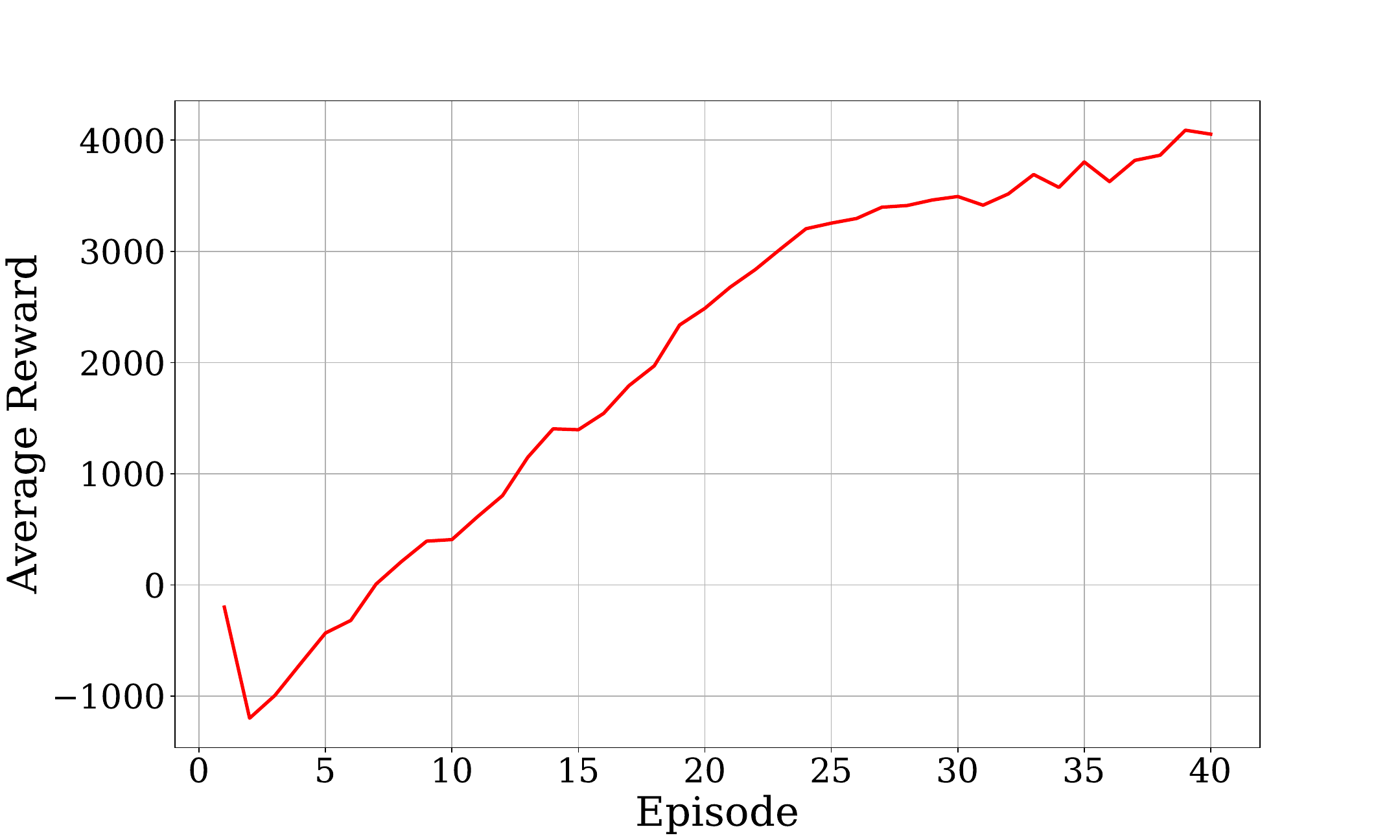}
    \caption{Driving Model}
    \label{fig:drive}
  \end{subfigure}
  \caption{Average Reward as a function of the training episode: the reward functions $R_b$ and $R_d$ for the braking model and driving model respectively}
  \label{fig:main}
\end{figure}

The driving DQN model takes $d$ and $\phi$ as inputs and outputs Q-values for five actions: going straight (1), turning left (2), turning right (3), turning slightly left (4), and turning slightly right (5). The model architecture consists of one fully connected layer with 8 units, followed by a ReLU activation, and a final output layer with 5 units, followed by a linear activation (Figure \ref{combined}). The model is trained on 40 episodes of alternating left turn and a right turn episodes, allowing the car to explore these two scenarios. The reward function given to the agent is designed to encourage the car to stay on the road and follow the planned trajectory while penalizing collisions. The reward function given to this model is as follow:

\begin{align*}
    R_d (d,\phi,a) = & R(s,\pi_d(s)) -10\,[|d| > 2] -200 \,[|d| > 3] -200 \,[|\phi| > 100] -200 \,[\textrm{collisions} > 0],
\end{align*}
with $R(s,\pi_d(s))$ a reward function that rewards the agent if it follows a predefined sub-optimal policy $\pi_d(s)$.

After training our agent, we observe that the average reward during training increases and reaches a plateau around 3700 (Figure \ref{fig:drive}). This shows that the car learned how to follow a path by staying on the road and avoiding collisions.

Now, we want to test the capacity of our model to generalize and create robust predictions. We test the driving model we got previously on 4 different trajectories in CARLA (Figure \ref{fig:traj}).

\begin{figure}[H]
\centering
\includegraphics[scale = 0.22]{ 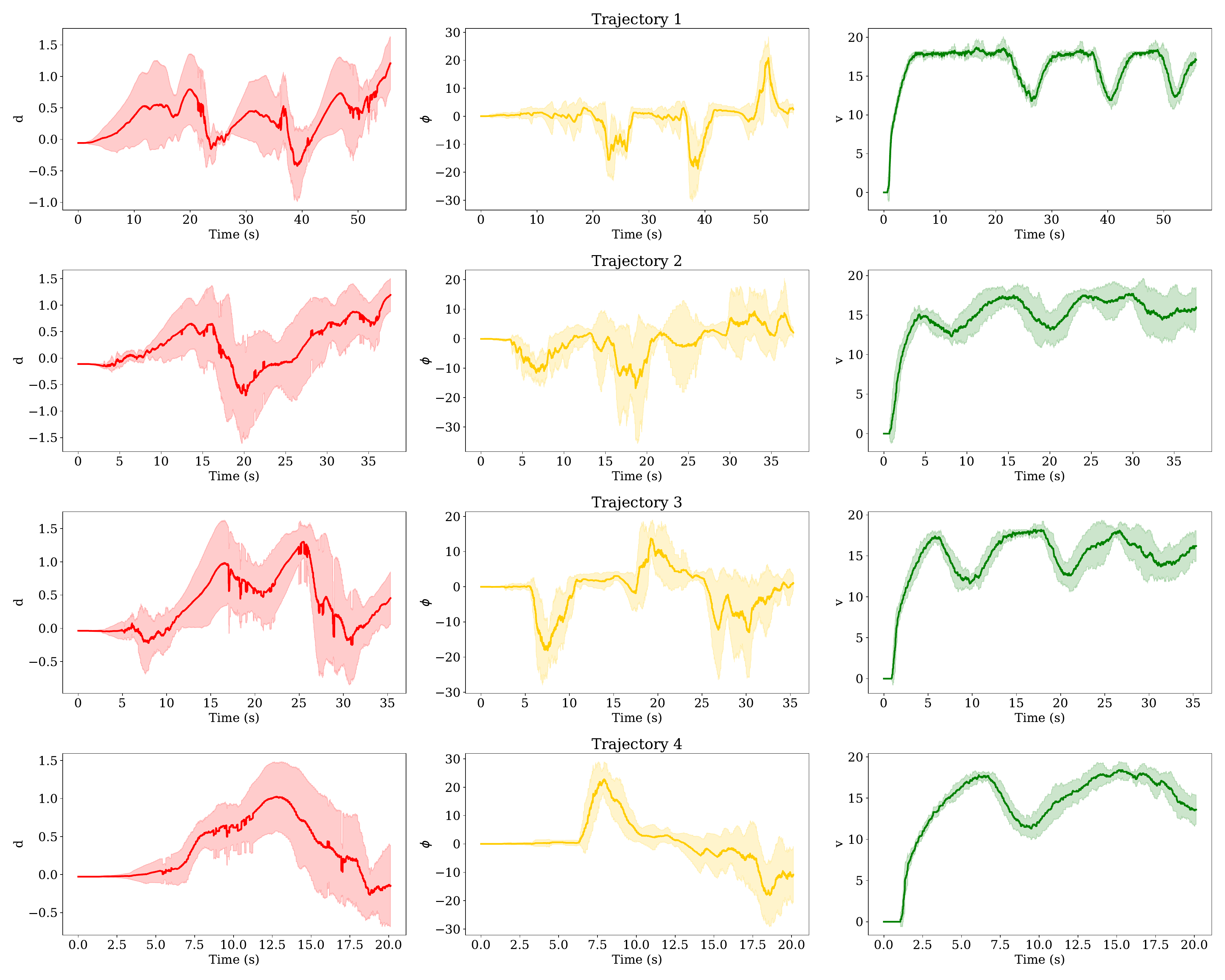}
\vspace{-0.3cm}
    \caption{The means and standard deviations of $d$, $\phi$ and $v$, achieved after the 10 trials, as a function of time  for each of the 4 trajectories}
\label{datatraj}
\end{figure}

We launch the agent 10 times on each trajectory and add 5\% of random actions to test the capacity of the model to correct its trajectory to remain on the road. We record the velocity $v$ of our agent, the difference in orientation angle with the road $\phi$, and the signed distance to the center of the road $d$. We obtain the plots in Figure \ref{datatraj}. We observe that for all the trajectories, the car reached its final destination while keeping $d$ smaller than 1.5 meters at all time, which means that the car was able to position itself on the road. Moreover, we can see that whenever the $|\phi|$ and $|d|$ values are increased, the car managed to take a sequence of actions that reduced them, and thus staying close on the planned path. Finally, we see that the agent slows down on right and left turns which shows that the car managed to learn some safety rules to minimize collisions and remain on the road.

\subsection{Integration of Braking and Driving Models}

\begin{figure}[H]
\centering
\includegraphics[scale = 0.2]{ 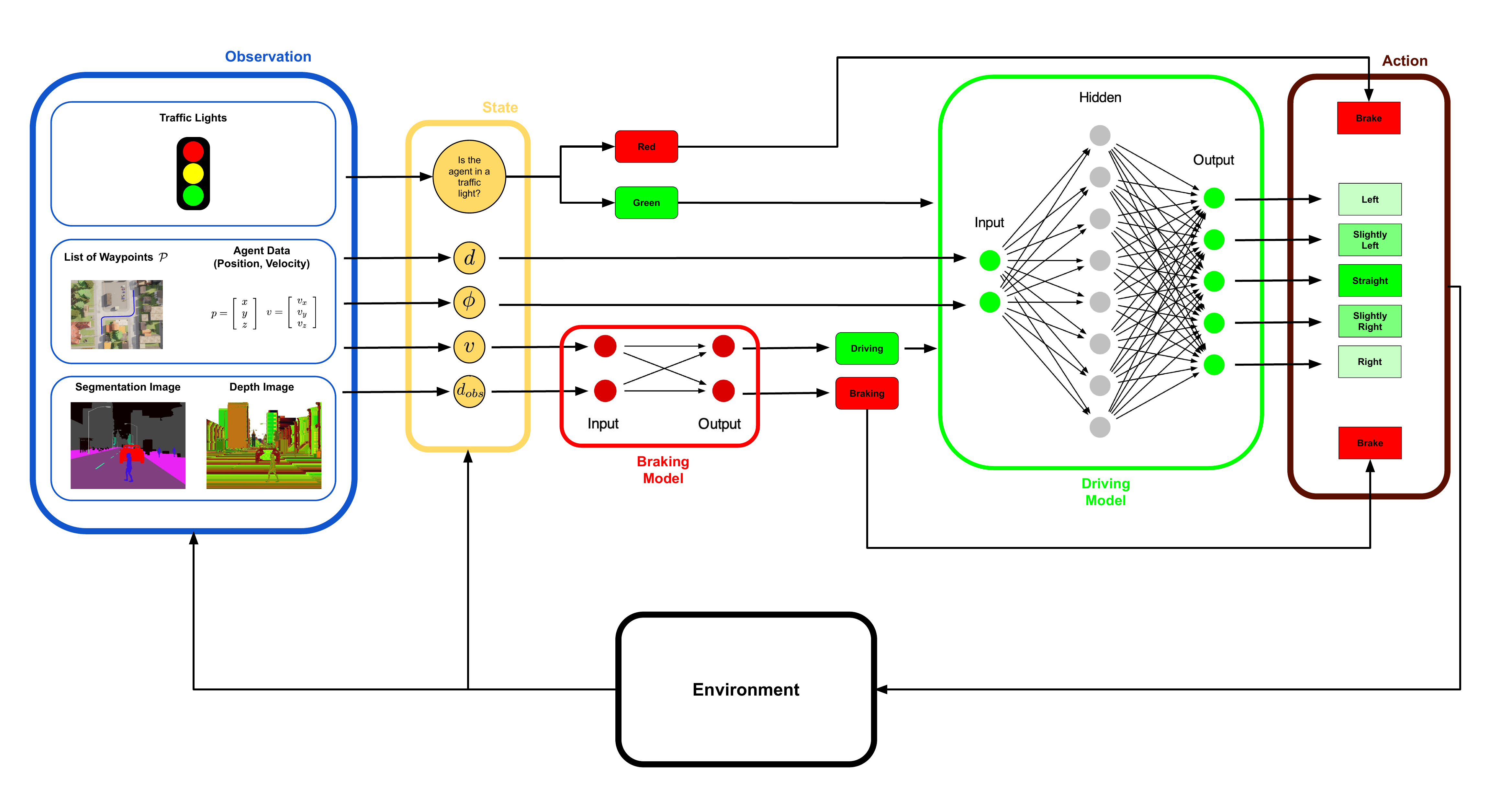}
    \caption{Final Architecture of our Combined Model to navigate the car safely in traffic}
\label{combined}
\end{figure}

To build a model that allows a car to drive safely and autonomously in traffic, we need to combine the braking and driving DQN models we obtained earlier. We use a hierarchical approach where the braking model is used as a safety net to prevent collisions when an obstacle is too close. The final model takes as input the values of $d$, $\phi$, $d_{obs}$, and $v$, obtained from the observations of the agent (Figure \ref{combined}). Before making predictions on which action to take, the agent checks if there is a traffic light. If it is red, then, it brakes. If the light is green, the braking model uses $v$ and $d_{obs}$ to make a prediction on whether it is safe to drive or if the car should brake. If it is safe to drive, the driving model uses $d$ and $\phi$ to make a prediction on which driving action to take (Figure \ref{combined}). 

This approach ensures that the car can safely navigate through traffic while following its path and reaching its final destination.

\section{Case Scenarios and Results}

The trained DQN models are tested on 4 different trajectories in a traffic scenario containing 35 other vehicles and 80 pedestrians. The model is trained on a different set of trajectories and below selected trajectories demonstrate the effectiveness of our models to generalize on unseen data (refer to trajectories from Figure \ref{fig:traj}). Table \ref{table1} shows the percentage of success and failure cases for each trajectory when tested for 25 runs each. 

\begin{table}[H]
  \centering
  
  \label{tab:trajectory-stats}
  \begin{tabular}{lcccccc}
    \toprule
    Trajectory & Path & Average & \% Success & \% of Collision & \% of Collision & Deadlocks \\
    Name & Distance (m) & Time (s) &  & with Vehicles & with Pedestrians &   \\
    \midrule
    Trajectory 1 & 258 & 321.5 & 96 & 0 & 4 & 0 \\
    Trajectory 2 & 163 & 177.4 & 92 & 4 & 0 & 4 \\
    Trajectory 3 & 150 & 148.18 & 92 & 4 & 4 & 0 \\
    Trajectory 4 & 104 & 136.48 & 96 & 0 & 4 & 0 \\
    Average & 168.75 & 195.89 & 94 & 2 & 3 & 1 \\
    \bottomrule
  \end{tabular}
  \vspace{0.5cm}
  \caption{Trajectory Statistics}
  \label{table1}
\end{table}

A run is considered to be successful if the car safely navigates to the final destination without colliding with an obstacle (vehicle, pedestrian, and sidewalk). Successful runs for trajectories 1, 2, 3, and 4 are shown in \href{https://drive.google.com/drive/folders/1dLaCMtb7UOpxs6karu0RshLH0Opd3Izp?usp=sharing}{Videos 4, 5, 6, and 7} respectively. We achieved a success rate of 94\% when we ran our model on 4 different trajectories for a total of 100 runs. 

In order to improve our success rate, we decide to investigate failure cases. We look at what may have caused those incidents in order to know the limitations of our model. 

\textbf{Reasons for failure cases:}

\begin{enumerate}
    \item [\textbullet]\textbf{Collision with Vehicle} - Evident from \href{https://drive.google.com/drive/folders/1dLaCMtb7UOpxs6karu0RshLH0Opd3Izp?usp=sharing}{Video 1}, an FoV of 40 is causing the agent to not see the car which is present right in front of it after turning. But as soon as it turns, it detects the car and starts braking but doesn’t stop on time.

    \item [\textbullet] \textbf{Collision with Pedestrian} - A low FoV is causing the agent to take an action of throttle instead of the brake as the distance from the closest object returned is 0. Evident from \href{https://drive.google.com/drive/folders/1dLaCMtb7UOpxs6karu0RshLH0Opd3Izp?usp=sharing}{Video 2}, for the car, due to limited FoV and opposite lane removal pre-processing step, the pedestrian is suddenly coming in front of it and cannot brake on time.

    \item [\textbullet] \textbf{Deadlock} - A deadlock typically happens at the intersection when the traffic cars get built up as they wait for the pedestrians to cross the road. As evident from \href{https://drive.google.com/drive/folders/1dLaCMtb7UOpxs6karu0RshLH0Opd3Izp?usp=sharing}{Video 3}, our model is not able to distinguish between an obstacle that is not blocking its intended path and one that is less than 100 units away in our distance map. 
\end{enumerate}

\section{Conclusion}

In this paper, we developed a robust model capable of generalizing well on unseen trajectories in traffic scenarios by taking input from only 4 variables and successfully reaching the final destination without any collisions. Passing the output of the camera sensors directly to the DQN models resulted in very poor generalization on unseen trajectories due to large state space.

And thus, it is necessary to pre-process the output data from the sensors and extract only the useful information. This significantly reduced the state space and the training efficiency and increased the model's capabilities to generalize well on unseen scenarios. Having trained the braking and driving models on short trajectories for 40 episodes each and getting very good results confirms the effectiveness of our approach. 

The next steps going forward would be to train a model by passing segmentation images as input and predict the values of $d$ and $\phi$, as getting waypoints in the real world is not possible.

The code is available here: \href{https://github.com/Tejas-Deo/Safe-Navigation-Training-Autonomous-Vehicles-using-Deep-Reinforcement-Learning-in-CARLA}{GitHub Repo Link}








\bibliographystyle{unsrt}
\bibliography{references}







\end{document}